\documentclass[10pt,twocolumn,letterpaper]{article}

\usepackage{cvpr}
\usepackage{times}
\usepackage{epsfig}
\usepackage{graphicx}
\usepackage{amsmath}
\usepackage{amssymb}
\usepackage{multirow}
\usepackage{color}
\usepackage{authblk}

\usepackage[pagebackref=true,breaklinks=true,letterpaper=true,colorlinks,bookmarks=false]{hyperref}

\cvprfinalcopy 

\def\cvprPaperID{2842} 

\ifcvprfinal\fi
\begin{document}

\title{Joint Sequence Learning and Cross-Modality Convolution for 3D Biomedical Segmentation}

\author[1]{Kuan-Lun Tseng}
\author[2]{Yen-Liang Lin}
\author[1]{Winston Hsu}
\author[1]{Chung-Yang Huang}
\affil[1]{National Taiwan University, Taipei, Taiwan}
\affil[2]{GE Global Research, NY, USA}
\affil[ ]{\tt\small\url{tkuanlun350@gmail.com,yenlianglintw@gmail.com, whsu@ntu.edu.tw, cyhuang@ntu.edu.tw}}
\renewcommand\Authands{ and }


\maketitle
\begin{abstract}
Deep learning models such as convolutional neural network have been widely used in 3D biomedical segmentation and achieve state-of-the-art performance.
However, most of them often adapt a single modality or stack multiple modalities as different input channels.
To better leverage the multi-modalities, we propose a deep encoder-decoder structure with cross-modality convolution layers to incorporate different modalities of MRI data.
In addition, we exploit convolutional LSTM to model a sequence of 2D slices, and jointly learn the multi-modalities and convolutional LSTM in an end-to-end manner.
To avoid converging to the certain labels, we adopt a re-weighting scheme and two-phase training to handle the label imbalance.
Experimental results on BRATS-2015 \cite{info:doi/10.2196/jmir.2930} show that our method outperforms state-of-the-art biomedical segmentation approaches.
\end{abstract}


\section{Introduction}

3D image segmentation plays a vital role in biomedical analysis.
Brain tumors like gliomas and glioblastomas have different kinds of shapes, and can appear anywhere in the brain, which make it challenging to localize the tumors precisely.
%
%
%
%
%
Four different modalities of magnetic resonance imaging (MRI) image are commonly referenced for the brain tumor surgery: T1 (spin-lattice relaxation), T1C (T1-contrasted), T2 (spin-spin relaxation), and FLAIR (fluid attenuation inversion recovery).
Each modality has distinct responses for different tumor tissues.
We leverage multiple modalities to automatically discriminate the tumor tissues for assisting the doctors in their treatment planning.

\begin{figure*}[t]
\graphicspath{{fig/}}
\begin{center}
\includegraphics[height=0.45\linewidth]{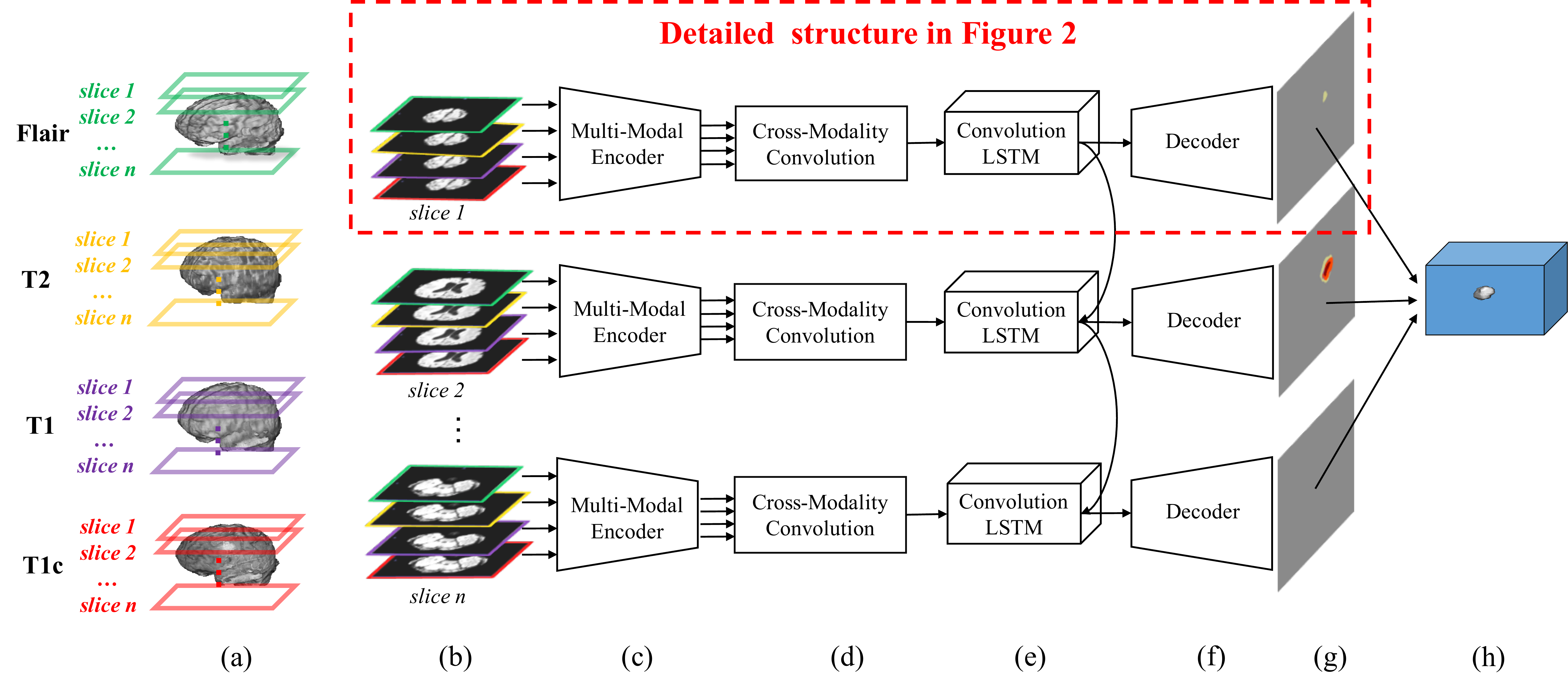}
\end{center}
\caption{
System overview of proposed method for 3D biomedical segmentation.
(a) We extract the slices (according to the depth values) from different modalities (i.e., Flair, T2, T1, T1c) of a 3D MRI image.
(b) The slices of the same depth from different modalities are combined together.
(c) Each slice in a stack is fed into the multi-modal encoder for learning a latent semantic feature representation.
(d) A cross-modality convolution is utilized to aggregate the information from different modalities.
(e) We leverage convolutional LSTM to better model the correlations between consecutive slices.
(f) A decoder network is used to up-sample the outputs of convolutional LSTM to the original resolution as the input slices.
(g) The final results (i.e., different types of tumor issues) are predicted at each pixel location.
(h) We stack the sequence of 2D prediction results into a 3D segmentation.
}
\label{fig:fig1}
\label{fig:onecol}
\end{figure*}
Recently, deep learning methods have been adopted in biomedical analysis and achieve the state-of-the-art performance.
%
%
%
%
Patch-based methods \cite{brats,havaei2016brain} extract small patches of an image (in a sliding window fashion) and predict the label for each central pixel.
These methods suffer from slow training, as the features from the overlapped patches are re-computed.
Besides, they only take a small region into a network, which ignores the global structure information (\eg, image contents and label correlations).
Some methods apply 2D segmentation to 3D biomedical data \cite{UNet} \cite{chen2016dcan}.
They slice a 3D medical image into several 2D planes and apply 2D segmentation for each 2D plane.
3D segmentation is then generated by concatenating the 2D results.
However, these 2D approaches ignore the sequential information between consecutive slices.
For examples, there may have rapid shape changes in the consecutive depths.
3D-based approaches \cite{lai2015deep} use 3D convolution to exploit different views of a 3D image.
But, they often require a larger number of parameters and are prone to overfitting on the small training dataset.
The above methods often stack modalities as different input channels for deep learning models, which do not explicitly consider the correlations between different modalities.
To address these problems, we propose a new deep encoder-decoder structure that incorporates spatial and sequential information between slices, and leverage the responses from multiple modalities for 3D biomedical segmentation.

Figure \ref{fig:fig1} shows the system overview of our method.
Given a sequence of slices of multi-modal MRI data, our method accurately predicts the different types of tumor issues for each pixel.
Our model consists of three main parts: multi-modal encoder, cross-modality convolution and convolutional LSTM.
%
%
%
The slices from different modalities are stacked together by the depth values (b).
Then, they pass through different CNNs in the multi-modal encoder (each CNN is applied to a different modality) to obtain a semantic latent feature representation (c).
Latent features from multiple modalities are effectively aggregated by the proposed cross-modality convolution layer (d).
Then, we leverage convolutional LSTM to better exploit the spatial and sequential correlations of consecutive slices (e).
A 3D image segmentation is generated (h) by concatenating a sequence of 2D prediction results (g).
Our model jointly optimizes the slice sequence learning and multi-modality fusion in an end-to-end manner.
The main contributions of this paper are summarized as following:
\begin{itemize}
\item We propose an end-to-end deep encoder-decoder network for 3D biomedical segmentation. Experimental results demonstrate that we outperform state-of-the-art 3D biomedical segmentation methods.
\item We propose a new cross-modality convolution to effectively aggregate the multiple resolutions and modalities of MRI images.
\item We leverage convolution LSTM to model the spatial and sequential correlations between slices, and jointly optimize the multi-modal fusion and convolution LSTM in an end-to-end manner.
\end{itemize}
%

\section{Related Work}
{\bf Image Semantic Segmentation.}
%
%
Various deep methods have been developed and achieve significant progress in image segmentation \cite{long2015fully, badrinarayanan2015segnet,chen2014semantic,noh2015learning}.
%
%
%
%
%
%
These methods use convolution neural network (CNN) to extract deep representations and up-sample the low-resolution feature maps to produce the dense prediction results.
%
%
SegNet \cite{badrinarayanan2015segnet} adopts an encoder-decoder structure to further improve the performance while requiring fewer model parameters.
%
%
%
We adopt the encoder-decoder structure for 3D biomedical segmentation and further incorporate cross-modality convolution and convolutional LSTM to better exploit the multi-modal data and sequential information for consecutive slices.

%
%
%
%
%
%
%

{\bf 3D Biomedical Image Segmentation.}
There have been much research work that adopts deep methods for biomedical segmentation.
Havaei et al. \cite{havaei2016brain} split 3D MRI data into 2D slices and crop small patches at 2D planes.
They combine the results from different-sized patches and stack multiple modalities as different channels for the label prediction.
Some methods utilize full convolutional network (FCN) structure \cite{long2015fully} for 3D biomedical image segmentation.
U-Net \cite{UNet} consists of a contracting path that contains multiple convolutions for downsampling, and a expansive path
that has several deconvolution layers to up-sample the features and concatenate the cropped feature maps from the contracting path.
However, depth information is ignored by these 2D-based approaches.

To better use the depth information, Lai et al. \cite{lai2015deep} utilize 3D convolution to model the correlations between slices.
However, 3D convolution network often requires a larger number of parameters and is prone to overfitting on small dataset.
kU-Net \cite{kUNet} is the most related to our work.
They adopt U-Net as their encoder and decoder and use recurrent neural network (RNN) to capture the temporal information.
Different from kU-Net, we further propose a cross-modality convolution to better combine the information from multi-modal MRI data, and jointly optimize the slice sequence learning and cross-modality convolution in an end-to-end manner.

{\bf Multi-Modal Images.}
In brain tumor segmentation, multi-modal images are used to identify the boundaries between the tumor, edema and normal brain tissue.
Cai et al. \cite{cai2007probabilistic} combine MRI images with diffusion tensor imaging data to create an integrated
multi-modality profile for brain tumors.
Their brain tissue classification framework incorporates intensities from each modality into an appearance signature of each voxel to train the classifiers.
Menze et al. \cite{menze2010generative} propose a generative probabilistic model for reflecting the differences in tumor appearance across different modalities.
%
%
%
In the process of manual segmentation of a brain tumor, different modalities are often cross-checked to
better distinguish the different types of brain tissue.
For example, according to Menze et al. \cite{menze2015multimodal},
the edema is segmented primarily from T2 images and FLAIR is used to cross-check the extension of the edema.
Also, enhancing and non-enhancing structures are segmented by evaluating the hyper-intensities in T1C.

Existing CNN-based methods (e.g., \cite{brats,havaei2016brain}) often treat modalities as different channels in the input data.
However, the correlations between them are not well utilized.
To our best knowledge, we are the first to jointly exploit the correlations between different modalities, and the spatial and sequential dependencies for consecutive slices.

\begin{figure*}[t]
\graphicspath{{fig/}}
\begin{center}
\includegraphics[height=0.5\linewidth]{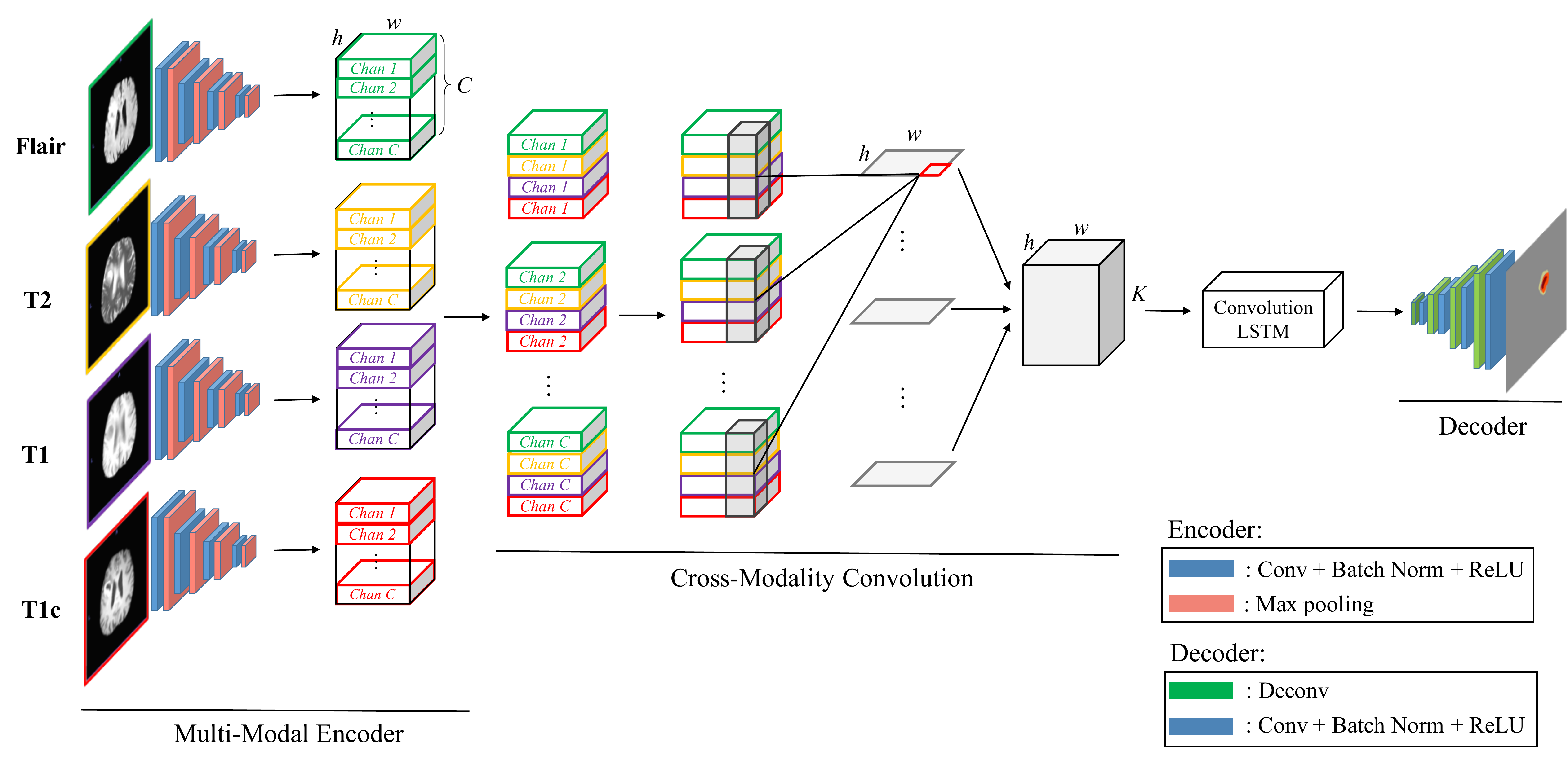}
\end{center}
\caption{
Cross-modality convolution is a weighting function across four different modalities.
The features extracted by the multi-modal encoder have the shape size $4 \times h \times w \times C$.
We reshape the input features to $C \times h \times w \times 4$ and apply cross-modality convolution with kernel size $4 \times 1 \times 1$  to produce a set of feature maps.
The feature maps are fed into convolutional LSTM and decoder to generate the final dense predictions.
}
\label{fig:fusion}
\end{figure*}
\section{Method}
Our method is composed of three parts, multi-modal encoder and decoder, cross-modality convolution, and convolution LSTM.
Encoder is used for extracting the deep representation of each modality. Decoder up-samples the feature maps to the original resolution for predicting the dense results.
Cross-modality convolution performs 3D convolution to effectively combine the information from different modalities.
%
%
Finally, convolutional LSTM further exploits the sequential information between consecutive slices.
\subsection{Encoder and Decoder}
Due to the small size of BRATS-2015 training dataset \cite{info:doi/10.2196/jmir.2930}, we want the parameter space of our multi-modal encoder and decoder to be relatively small for avoiding the overfitting.
Figure \ref{fig:fusion} shows our multi-modal encoder and decoder structure.
We adopt the similar architecture as in SegNet \cite{badrinarayanan2015segnet} for our encoder, which comprises four convolution layers and four max pooling layers.
Each convolution layer uses the kernel size $3\times3$ to produce a set of feature maps, which are further applied by a batch normalization layer \cite{ioffe2015batch} and an element-wise rectified-linear non-linearity (ReLU).
%
%
%
Batch normalization layer is critical for training our network, as the distributions of tumor and non-tumor tissues can vary from one slice to another even in the same brain.
%
%
%
%
%
%
%
%
%
%
%
Then, a max pooling layer with size 2 and stride 2 is applied and
the output feature maps are down-sampled by a factor of 2.

For the decoder network,
each deconvolution layer performs the transposed convolution.
%
%
%
Then, a convolution and batch normalization are applied.
After up-sampling the feature maps to the original resolution,
we pass the output of the decoder to a multi-class soft-max classifier to produce the class probabilities of each pixel.
\subsection{Multi-Resolution Fusion (MRF)}
\begin{figure}[t]
\graphicspath{{fig/}}
\begin{center}
\includegraphics[height=0.6\linewidth]{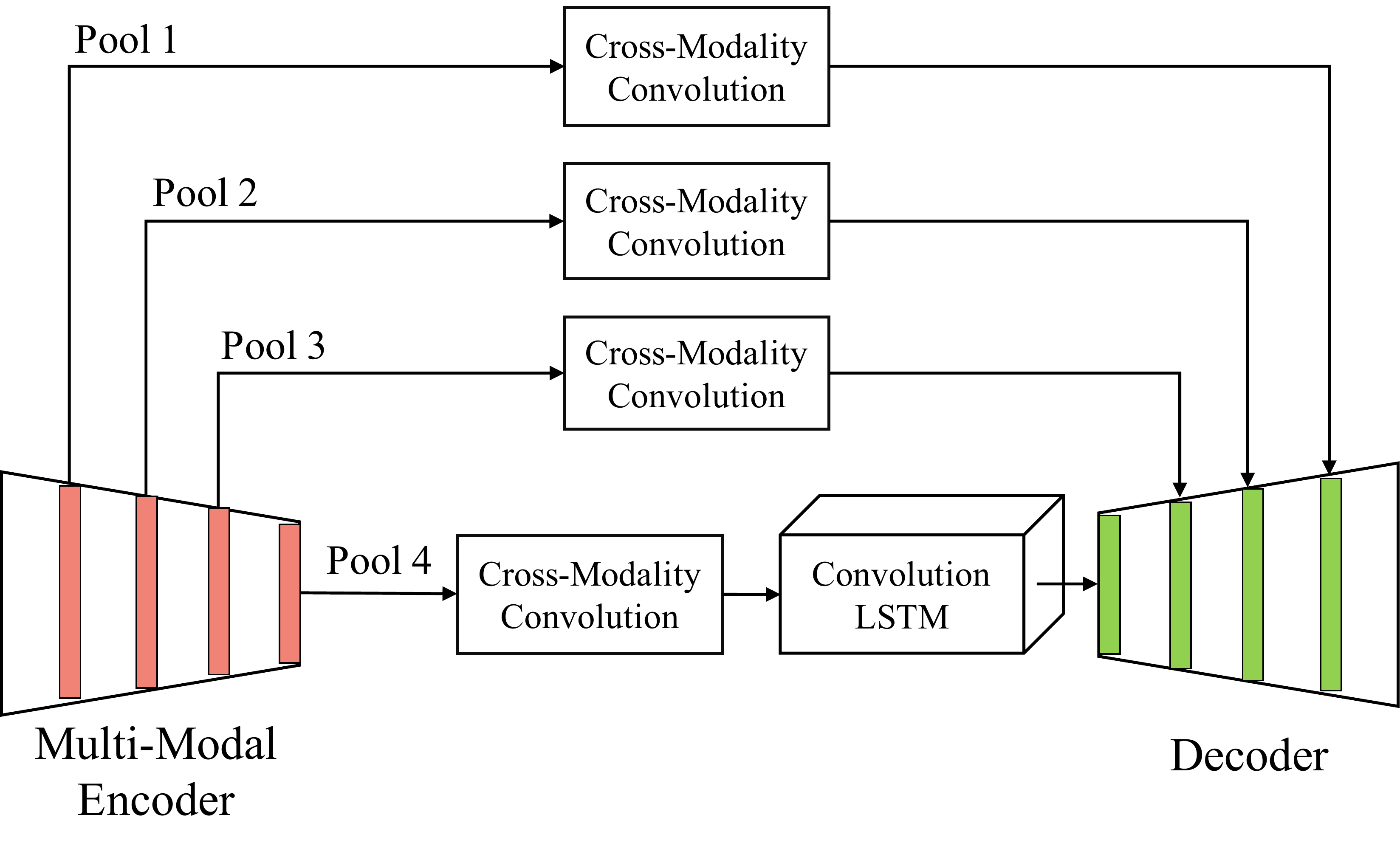}
\end{center}
\caption{
System overview of our multi-resolution fusion strategy.
In the multi-modal encoder, the feature maps generated after each pooling layer are applied with
cross-modality convolution to aggregate the information between modalities.
Following that, those feature maps are multiplied with the up-sampled feature maps from the decoder to combine the multi-resolution information.
}
\label{fig:fig2}
\label{fig:onecol}
\end{figure}

Recent image segmentation models
\cite{eigen2015predicting,long2015fully,hariharan2015hypercolumns,UNet}
fuse multi-resolution feature maps with the concatenation.
Feature concatenation often requires additional learnable weights because of the increase of channel size.
In our method, we use the feature multiplication instead of concatenation.
The multiplication does not increase feature map size and therefore no additional weights are required.
%
%
%
We combine the feature maps from the encoder and decoder networks, and train the whole network end-to-end.
The overall structure of cross-modality convolution (CMC) and multi-resolution fusion are shown in Figure~\ref{fig:fig2}.
We perform CMC after each pooling layer in the multi-modal encoder, and multiply it with the up-sampled feature maps from the decoder
to combine multi-resolution and multi-modality information.
We will explain the details of CMC layer in the next section.

\subsection{Cross-Modality Convolution (CMC)}
%
%
%
We propose a cross-modality convolution (CMC) to aggregate the responses from all modalities.
After the multi-modal encoder, each modality is encoded to a feature map of size $h \times w \times C$, where $w$ and $h$ are feature dimensions, and $C$ is the number of channels.
%
%
%
We stack the features of the same channels from four modalities into one stack.
After reshaping, we have $C \times 4 \times h \times w$ feature maps.
Our cross-modality convolution performs 3D convolution with the kernel size $4 \times 1 \times 1$, where 4 is the number of modalities.
As the 3D kernel convolves across different stacks, it assigns different weights to each modality and sums the feature values in the output feature maps.
The proposed cross-modality convolution combines the spatial information of each feature map and models the correlations between different modalities.
%

%
%
%
%
%
%
%
\subsection{Slice Sequence Learning}
%
%
We propose an end-to-end slice sequence learning architecture to capture the sequential dependencies.
We use image depths as a sequence of slices and leverage convolutional LSTM \cite{xingjian2015convolutional} (convLSTM) to model the slice dependencies.
Different from traditional LSTM \cite{hochreiter1997long}, convLSTM replaces the matrix multiplication by the convolution
operators in state-to-state and input-to-state transitions, which preservers the spatial information for long-term sequences.
%
%

{\bf Convolutional LSTM (convLSTM).}
The mechanism is similar to the traditional LSTM except that we replace the matrix multiplication by a convolution operator ${*}$.
The overall network is defined as following:
\[\begin{array}{l}
{i_t} = \sigma ({x_t}*{W_{xi}} + {h_{t - 1}}*{W_{hi}} + {b_i})\\
{f_t} = \sigma ({x_t}*{W_{xf}} + {h_{t - 1}}*{W_{hf}} + {b_f})\\
{c_t} = {c_{t - 1}} \circ {f_t} + {i_t} \circ \tanh ({x_t}*{W_{xc}} + {h_{t - 1}}*{W_{hc}} + {b_c})\\
{o_t} = \sigma ({x_t}*{W_{xo}} + {h_{t - 1}}*{W_{ho}} + {b_o})\\
{h_t} = {o_t} \circ \tanh ({c_t})
\end{array}\]
Where $\sigma$ is a sigmoid function and $\tanh$ is a hyperbolic tangent function.
There are three gates, namely input gate ${i_t}$, forget gate ${f_t}$ and output gate ${o_t}$.
The forget gate controls whether to remember previous cell state ${c_{t-1}}$ by the output of activation function $\sigma$.
Similarly, input gate controls whether new candidate value should be added into new cell state ${c_t}$.
Finally, the output gates controls which parts we want to produce.
The output size of feature maps depends on the kernel size and padding methods.
%
%

Our slice sequence learning architecture combines the encoder-decoder network with a convLSTM to better model a sequence of slices.
ConvLSTM takes a sequence of consecutive brain slices encoded by multi-modal encoder and CMC (Figure~\ref{fig:fig1} (e)).
%
%
%
The weights in convLSTM are shared for different slices, therefore the parameter size does not increase linearly as the length of slice sequence growths.
The output feature maps of convLSTM are up-sampled by the decoder network (Figure~\ref{fig:fig1} (f)).
%

\section{Experiments}
We conduct experiments on two datasets to show the utility of our model.
We compare our cross-modality convolution with the traditional methods that stack modalities as different channels.
We evaluate our sequence learning scheme on typical video datasets to verify our method for modeling the temporal dependency.
We also evaluate our methods on a 3D biomedical image segmentation dataset.
\subsection{Dataset}
{\bf CamVid dataset.}
Cambridge-driving labelled video database (CamVid) dataset \cite{BrostowSFC:ECCV08}
is captured from the perspective of a driving automobile with fixed-position CCTV-style cameras.
CamVid provides videos with object class semantic labels.
The database provides ground truth labels that associate each pixel with one of 32 semantic classes.
We split the CamVid dataset with 367 training, 100 validation and 233 testing images.
%
The evaluation criteria is the mean intersection over union (Mean IU).
Mean IU is a commonly-used segmentation performance measurement that calculates the ratio of the area of intersection to the area of unions.
%
%
Selected images are sampled from the original videos and down-sampled to $640\times480$.
The length between two consecutive image is 30 frames long.

{\bf BRATS-2015 dataset.}
BRATS-2015 training dataset comprises of 220 subjects with high grade gliomas and 54 subjects with
low grade gliomas.
The size of each MRI image is $155\times240\times240$.
We use 195 high grade gliomas and
49 low grade gliomas for training, and the rest 30 subjects for testing.
We also conduct five-fold evaluation by using BRATS-2015 online judge system for avoiding overfitting.
All brain in the dataset have the same orientation and the four modalities are synchronized.
The label image contains five labels: non-tumor, necrosis, edema, non-enhancing tumor and enhancing tumor.
The evaluation system separates the tumor structure into three regions due to practical clinical applications.
\begin{itemize}
\item Complete score: it considers all tumor areas and evaluates all labels 1, 2, 3, 4 (0 for normal tissue, 1 for edema, 2 for non-enhancing core, 3 for necrotic core, and 4 for enhancing core).
\item Core score: it only takes tumor core region into account and measures the labels 1, 3, 4.
\item Enhancing score: it represents the active tumor region, i.e., only containing the enhancing core (label 4) structures for high-grade cases.
\end{itemize}
There are three kinds of evaluation criteria: Dice, Positive Predicted Value and Sensitivity.
\[
Dice = \frac{P_1 \bigcap T_1}{(P_1 + T_1)/2}
\]
\[
PPV = \frac{P_1 \bigcap T_1}{P_1}
\]
\[
Sensitivity = \frac{P_1 \bigcap T_1}{T_1},
\]
where T is ground truth label and P is predicted result.
$T_1$ is the true lesion area and $P_1$ is the subset of voxels predicted as positives for the tumor region.
%

%
%
%
%
%
%
%
%
%
%
\begin{table*}[]
\centering
\begin{tabular}{l|l|l|l|l|l|l}
Method          								& label 0 & label 1 & label 2 & label 3 & label 4 & MeanIU 				\\ \hline
U-Net \cite{UNet}          						& 92.3    & 42.9    & 73.6    & 45.3    & 62.0    & 54.2   				\\
U-Net + two phase 							& 98.6    & 43.8    & 67.4    & 24.0    & 60.5    & 59.3   				\\ \hline
MME + MRF + CMC       						& 98.2    & 47.0    & 72.2    & 41.0    & 72.2    & 61.8   				\\
MME + MRF + CMC + two-phase   				& {\bf99.1}    & 48.8    & 63.8    & 36.2    & 76.9    & 64.0   				\\
MME + MRF + CMC + convLSTM           			& 96.6    & {\bf94.3}    & 71.2    & 32.8    & {\bf96.0}    & 62.5   			\\
MME + MRF + CMC + convLSTM + two-phase  	& 98.5    & 92.1    & {\bf77.3}    & {\bf55.9}    & 78.6    & {\bf73.5}   		\\
\end{tabular}
\caption{
MeanIU on our BRATS-2015 testing set with 30 unseen patients.
The results show that the proposed model: MME + MRF + CMC and MME + MRF + CMC + convLSTM achieve the best performance and outperform the baseline method U-Net \cite{UNet}.
Two-phase training better handles the label imbalance problem and significantly improves the results.
}
\label{t1}
\end{table*}

\begin{table}[]
\centering
\begin{tabular}{l|llll}
Method                                 					& MeanIU &  &  &  \\ \cline{1-2}
SegNet \cite{badrinarayanan2015segnet}                   & 47.85  &  &  &  \\
Encoder-Decoder + convLSTM                         		& 48.16  &  &  &  \\
Encoder-Decoder + MRF         					& 49.91  &  &  &  \\
Encoder-Decoder + convLSTM + MRF 			& {\bf 52.13}  &  &  &
\end{tabular}
\caption{
MeanIU on CamVid test set. Our encoder-decoder model with convolutional LSTM and multi-resolution fusion achieve the best results.
}
\label{t3}
\end{table}

\begin{table}[]
\centering
\begin{tabular}{l|llll}
Method                               		 			& MeanIU &  &  &  \\ \cline{1-2}
U-Net \cite{UNet}                    					& 54.3  &  &  &  \\
U-Net+two-phase                     					& 59.3  &  &  &  \\ \cline{1-2}
Encoder-Decoder                                			& 44.14  &  &  &  \\
MME + CMC                 						& 45.80  &  &  &  \\
Encoder-Decoder + MRF          					& 55.37  &  &  &  \\
MME + CMC + MRF 							& 61.83  &  &  &  \\
MME + CMC + MRF + two-phase    				& 64.02  &  &  &  \\
MME + CMC + MRF + convLSTM              		& 62.15  &  &  &  \\
MME + CMC + MRF + convLSTM + two phase    	& {\bf 73.52}  &  &  &
\end{tabular}
\caption{Segmentation results of our models on BRATS-2015 testing set with 30 unseen patients.}
\label{t4}
\end{table}

\subsection{Training}
{\bf Single Slice Training.}
The critical problem in training a fully convolutional network in BRATS-2015 dataset is that the label distribution
is highly imbalanced.
Therefore, the model easily converges into local minimum, i.e., predicting every pixel as non-tumor tissue.
We use \textit{median frequency balancing} \cite{eigen2015predicting} for handling the data imbalance, where the weight assigned to a class in the cross-entropy loss function is defined as:
\[{\alpha _c} = median\_freq/freq(c)\]
where $freq(c)$ is the number of pixels of class $c$ divided by the total number of pixels in
images where $c$ is present, and $median\_freq$ is the median of all class frequencies.
Therefore the dominant labels will be assigned with the lowest weight which balances the training process.

In our experiments, we find that weighted loss formulation will overly suppress the learning of dominant labels (e.g., label 0 for normal tissues) and wrongly predict the labels.
%
%
%
%

{\bf Two-Phase Training.}
In the first phase, we only sample the slices that contain tumor tissues and use \textit{median frequency balancing} method for de-weighting the losses of the dominant classes (e.g., normal tissue and background).
%
%
In the second phase, we replace the \textit{median frequence balancing} and use a lower learning rate (\ie,  10e-6 in our experiments) for training the model.
With the true distribution reflected on loss function, we can train our model in a more balanced way to preserve diversity and
adjust the output prediction to be closer to the real distribution of our data.
Two-phase training alleviates the problem of overly suppressing the dominant classes and learn much better results.

{\bf Slice Sequence Training.}
%
%
We avoid sampling the empty sequences (all the slices in the sequence are normal brain tissues) in the first training phase
to prevent the model from getting trapped into local minimum and apply two-phase training scheme for slice sequence leanring.
%
%
%
%
%
%

For training the convolutional LSTM, we adopt the orthogonal initialization \cite{saxe2013exact} for handling the missing gradient issue.
For CamVid dataset, we use the batch size of 5 and sequence length of 3.
For BRATS-2015 dataset, we use the batch size of 3 and sequence length of 3.
The initial learning rate is 10e-4 and we use Adam optimizer for training the network.

\begin{figure*}[t]
\graphicspath{{fig/}}
\begin{center}
\includegraphics[height=0.6\linewidth]{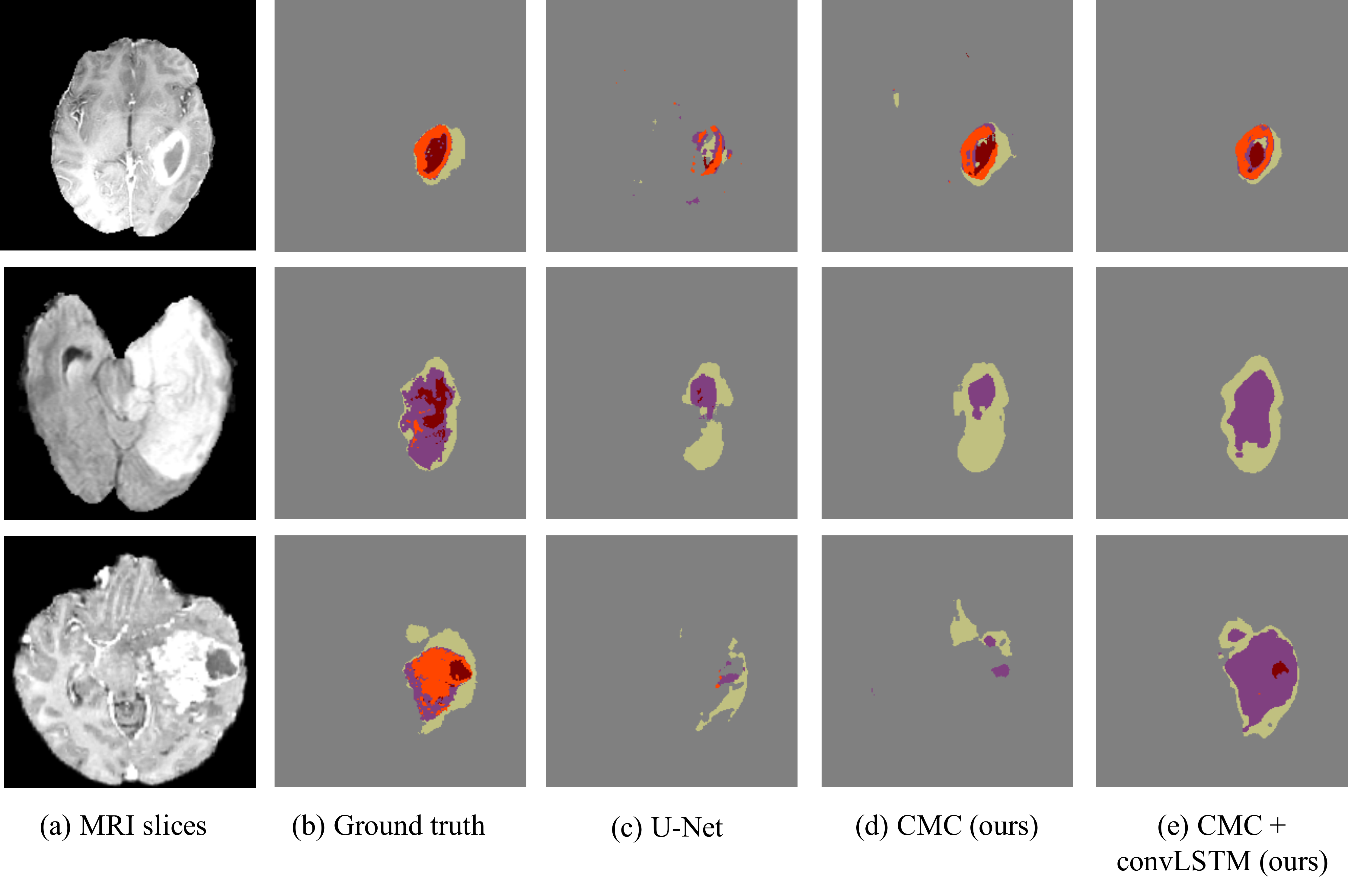}
\end{center}
\caption{
Segmentation results of variants of our method and U-Net \cite{UNet} baseline.
}
\label{fig:fig3}
\label{fig:onecol}
\end{figure*}


\subsection{Baseline}
The most relevant work to our method are kU-Net \cite{kUNet} and U-Net \cite{UNet}.
Both models achieve the state-of-the-art results in 3D biomedical image segmentation.
However, kU-Net is not originally designed for brain tumor segmentation, and the source code is not publicly available.
Therefore, we compare our method with U-Net, which shows competitive performance with kU-Net.
Original implementation of U-Net does not adopt batch normalization.
However, we find that it can not converge when training on BRATS-2015 dataset.
Thus, we re-implement their model with Tensorflow \cite{tensorflow2015-whitepaper} and incorporate batch normalization layer before every non-linearity in the contracting and expansive path of U-Net model.
We use orthogonal initialization and set the batch size to 10.
%
%
%
The inputs for U-Net is 4-channel MRI slices that stack four modalities into different channels.
We also investigate two-phase training and re-weighting for U-Net.
%
%
\begin{table*}[]
\centering
\begin{tabular}{|l|c|c|c|c|c|c|c|c|c|}
\hline
\multirow{2}{*}{Method} & \multicolumn{3}{c|}{Dice}                                                                  & \multicolumn{3}{c|}{Sensitivity}                                                               & \multicolumn{3}{c|}{PPV}                                                                   \\ \cline{2-10}
                        & \multicolumn{1}{l|}{Complete} & \multicolumn{1}{l|}{Core} & \multicolumn{1}{l|}{Enhancing} & \multicolumn{1}{l|}{Complete} & \multicolumn{1}{l|}{Core} & \multicolumn{1}{l|}{Enhancing} & \multicolumn{1}{l|}{Complete} & \multicolumn{1}{l|}{Core} & \multicolumn{1}{l|}{Enhancing} \\ \hline
U-Net \cite{UNet}                   	& 0.8504                          & 0.6174                      & 0.6793                          & 0.8727                          & 0.5296                      & 0.7229                           & 0.8376                          & 0.7876                      & 0.7082                           \\ \hline
Ours       					& 0.8522                          & 0.6835                      & 0.6877                           & 0.8741                          & 0.6545                     & 0.7735                           & 0.9117                         & 0.7945                      & 0.7212                          \\ \hline
\end{tabular}
\caption{
Five-fold cross validation segmentation results on 30 unseen patient evaluated by BRATS-2015 online system. Our model uses MME + MRF + CMC + convLSTM + two-phase settings, and outperforms U-Net \cite{UNet} in different measurements.}
\label{t2}
\end{table*}
\subsection{Experimental Results}
We conduct the experiments to evaluate cross-modality convolution (CMC).
We compare the performance of our multi-modal encoder and CMC layers with an encoder-decoder model (see Table~\ref{t4}).
The encoder-decoder model refers to a single encoder and decoder network without fusion.
The input of the encoder-decoder model stacks different modalities as different channels, while the input of our MME+CMC is illustrated in Figure~\ref{fig:fig1}(b).
%
%
The performance of our MME+CMC outperforms the basic encoder-decoder structure by approximately two percent in Mean IU.
%
%
Currently, the feature maps extracted by MME are down-sampled to a lower resolution, thus a certain amount of spatial information is lost.
We conjecture that the performance of our CMC could be further improved by incorporating higher resolution feature maps.

We conduct another experiment by using multi-resolution feature maps with CMC to verify whether multiple resolution helps.
%
%
%
%
In Table~\ref{t4}, we can observe that MRF significantly improves the performance, e.g.,
encoder-decoder with MRF improves the basic encoder-decoder by 10 percent.
We also evaluate our feature multiplication and feature concatenation used by U-Net, and find that they achieve similar performance.

Table~\ref{t1} shows that MME+CMC+MRF outperforms U-Net (similar to our encoder-decoder+MRF)
on Mean IU (almost every label except for label 0 (normal tissues)).
Because of the number of normal tissues are much larger than the other labels, the accuracy of label 0 is critical in Mean IU metric.
As a result, we use two-phase training to refine our final prediction.
After the second phase of training, the accuracy for label 0 is improved and our model shows much clean prediction results (cf. Figure~\ref{fig:fig3}).
%
%
%
%
%

%
To verify the generalizability of our sequence learning method, we further perform the slice-to-sequence experiments on CamVid dataset.
The sequence length used in the CamVid experiment is three and the settings for encoder-decoder are the same as BRATS dataset.
We incorporate convolutional LSTM with both basic encoder-decoder model and encoder-decoder+MRF.
Results show that convolutional LSTM consistently improves the performance for both settings (cf. Table~\ref{t3})
and outperforms SegNet \cite{badrinarayanan2015segnet}.
%
%
%
%
%
Due to the ability of convolutional LSTM for handling long-short term sequences and preserving the spatial information at the same time, the dependencies
between slices are well learned.

Our slice-to-sequence also improves the results of BRATS dataset.
In Table~\ref{t1}, we can see that the accuracy of label 1 to 4 (row 5 and 6) is much better than the single slice training (row 3 and 4).
After the second training phase, the accuracy of label 0 is improved and the model achieves the Mean IU 73.52, which outperforms single slice training model by a large margin.
%

%
%
In Table~\ref{t2}, we compare our slice-to-sequence model with U-Net on BRATS-2015 online system (we upload the five fold cross validation results to BRATS-2015 online evaluation system).
Two-phase training is applied to both methods and trained with the same epochs (without post-processing).
Our slice-to-sequence model outperforms U-Net in different measurements.
The visualized results also show that sequential information improves the predictions for detailed structures (cf. Figure~\ref{fig:fig3}).

Experimental results show that the proposed cross-modality convolution can effectively aggregate the
information between modalities and seamlessly work with multi-resolution fusion.
The two components are combined to achieve the best results.
The slice-to-sequence architecture further utilizes the sequential dependencies
to refine the segmentation results.
This end-to-end trainable architecture shows many potentials since
it provides consistent improvement on every configurations of our model.

\section{Conclusions}
In this paper, we introduce a new deep encoder-decoder architecture for 3D image biomedical segmentation.
We present a new cross-modality convolution to better exploit the multi-modalities, and a sequence learning method to integrate the information from consecutive slices.
We jointly optimizing sequence learning and cross-modality convolution in an end-to-end manner.
Experimental results on BRATS-2015 dataset demonstrate that our method improves the state-of-the-art methods.
\section{Acknowledgement}
This work was supported in part by the Ministry of Science and Technology, Taiwan, under Grant MOST 104-2622-8-002-002 and MOST 105-2218-E-002 -032, and in part by MediaTek Inc, and grants from NVIDIA and the NVIDIA DGX-1 AI Supercomputer.
{\small
\bibliographystyle{ieee}
\bibliography{egbib}
}

\end{document}